# A Corpus-based Analysis of Attitudinal Changes in Lin Yutang's Self-translation of *Between Tears and Laughter*

Zhiping Bai[a]*

[a]*Keyi College, Zhejiang Sci-tech University, Shaoxing, China*

BaiZhiping@ky.zstu.edu.cn

# A Corpus-based Analysis of Attitudinal Changes in Lin Yutang's Self-translation of *Between Tears and Laughter*

Attitude is omnipresent in almost every type of text. There has yet to be any relevant research on attitudinal shifts in self-translation. The Chinese version of *Between Tears and Laughter* is a rare case of self-translation and co-translation in that the first 11 chapters are self-translated by Lin Yutang, and the last 12 chapters by Xu Chengbin. The current study conducted a word frequency analysis of this book's English and Chinese versions with LIWC and AntConc, and made comparative research into Lin Yutang's attitudinal changes. The results show that due to different writing purposes and readerships, there is less anger in Lin's self-translation (*M*=0.7755, *SD*=0.2775) than in the first 11 chapters of the English original (*M*=1.1036, *SD*=0.3861), which is a significant difference (*t*=2.2892, *p*=0.0331). This attitudinal change is also reflected in the translations of some n-grams containing *anger* words. In contrast, there is no significant difference (*t*=1.88, *p*=0.07) between Xu's co-translation and the corresponding part of the original in attitude "anger". This paper believes that corpus tools can help co-translators keep their translation consistent in attitude.

**Introduction**

Lin Yutang (1895 – 1976), the author of *My Country and My People* and *Moment in Peking*, was a world-renowned Chinese writer publishing novels, essays, translations, textbooks, and Chinese-English dictionaries, making outstanding achievements in literature, translation, and language research. He even invented a Chinese typewriter. He wrote more than 30 books in English, most of which describe China's cultural aspects. According to the biography of Lin Yutang written by his daughter Lin Taiyi, Lin was nominated twice for the Nobel Prize for Literature (Lin 1989). In 1924, he published an essay named *Soliciting Translations of Essays and Advocating "Humor"* in *The Supplement to the Morning Newspaper* to be the first scholar to transliterate "humor" into "幽默", which is the only Chinese equivalent for "humor" now. Lin advocated witty essays, and his lacerating wit earned him the accolade "Master of Humor" from Chinese people.

Lin was born in the Fujian Province. Fujian's cultural exchanges with the West had already been frequent at that time. According to his *Memoirs of an Octogenarian* written in 1975, Lin became a Christian when he was a little boy because his father, Lin Zhicheng, was a pastor. The primary and secondary schools Lin attended were all missionary schools. Therefore, Lin had been influenced by Classical Chinese and Western cultures from an early age. After graduating from St. John's University in Shanghai, Lin began to teach English at Tsinghua University for three years before attending Harvard Graduate School of Arts & Sciences. Afterward, he completed his doctorate at Leipzig University and returned to China to join the faculty of Xiamen University. In 1936, one year after publishing *My Country and My People*, Lin went to live in America for over 20 years. As a bilingual writer, he knew so well about American culture that the Occident might have influenced his philosophy. In his book *On the Wisdom of America*, he appreciated the humanitarian insights of John Dewey, George Santayana, and William James. In his novels, we can easily see the combination of eastern and western elements. Just as Lin commented on himself in the first chapter of his autobiography, he was a "bundle of contradictions". In his book *From Pagan to Christian*, we can see how he changed his philosophical views in his life, from giving up Christianity in his middle age to returning to embracing it once more in his old age. Therefore, some attitudinal "contradictions" might exist between his English works and Chinese self-translations.

As a translator, he translated both English and Chinese works, such as Dora Russell's *Woman and Knowledge*, George Bernard Shaw's *Pygmalion*, and Qing

Dynasty author Shenfu's *Six Chapters of a Floating Life*. Lin was also one of the few self-translators in the translation history of China. The other three well-known self-translators are Zhang Ailing, Xiao Qian, and Bai Xianyong. Most of Lin's self-translations are essays. In the 1930s, Lin translated his English essays published in *The China Critic Weekly* and published them one or two years later in the Chinese Journals *Analects*, *This Human World*, and *The Cosmic Wind*, founded by himself. Qian Suoqiao (2012) compiled some of Lin's original articles and their corresponding self-translations in *The Little Critic: The Bilingual Essays of Lin Yutang*. Besides self-translations of essays, Lin only translated the first 11 chapters of his book, *Between Tears and Laughter*, and Xu Chengbin translated the last 12 chapters. Since he was too busy to translate his own books, all of the Chinese translations of his other English books were done by other translators. Although those translations were well-received, he was not always pleased with them (Lin 1989).

*Between Tears and Laughter* is a book on politics and philosophy. John Day Book Company published the English original in 1943. There is a stark difference between Lin's English and Chinese prefaces. The 111-word English preface's tone seems earnest and rational. In contrast, the Chinese preface is a new preface of 2328 Chinese characters and 8 English words, tailor-made for domestic readers. In it are criticisms of Occidental imperialism and power politics as well as encouragements for Chinese compatriots.

This book criticizes Western materialist philosophy in general and power politics in particular. However, the writing purpose of this book is not only to express dissatisfaction and criticize war politics and power politics but also to show excellent traditional Chinese cultural elements, especially the spirit of courtesy and accommodation, which might be the only possible basis for a more civilized world order. Lin pointed out that traditional Chinese philosophy might be the solution to world peace. For example, the philosophies of Confucius, Mencius, and Laotse can all contribute to the philosophy of world peace. Lin did not lose his hope of arousing sympathy and respect from the Western powers for China.

Prior research has proven that emotion is a tool for persuasion (Wegener and Petty 1994; Tannenbaum et al. 2015). There is a strong association between emotion and persuasion (Rocklage, Rucker, and Nordgren 2018). Outward displays of emotions evolved primarily to influence others (Frijda and Mesquita 1994). Hence, in order to achieve the writing purpose of international communication and publicity, this book must have emotional appeal, for which emotive emotional attitudes are essential. In light of this, Lin might have considered readers' acceptance and made changes to the specific contents of this book. However, contrary to his intent, the emotional appeals in the English version backfired and incurred criticisms from some book reviewers (e.g., Vinacke 1944).

Popovič (1975) defined "self-translation" as "the author's translation of his own original work". Self-translation does not distort the intentions of the original text, as is the case with ordinary translation. Panichelli-Batalla (2015) pointed out that self-translators who are also authors are generally considered to have greater freedom than other translators, with priority for revision and improvement of the original text. Lin chose to self-translate this book to encourage his compatriots in the shortest possible time. Since there are many descriptions of the then-current affairs, Lin collaborated with a co-translator, Xu Chengbin, to speed up the translation and soon published the Chinese version in 1944. After completing the translation, Lin immediately visited Chongqing and other major cities to give speeches. He hoped his patriotic and peace-loving enthusiasm would be understood and supported. Therefore, Lin might have modified particular negative descriptions in the English original to make the translation more acceptable.

Attitudes continue to be one of the essential concepts of social psychology, and attitude research is one of its most active areas (Bohner and Dickel 2011). A person's values, general goals, language, emotions, life span, and developmental aspects influence their attitudes and persuasion processes (Albarracin and Shavitt 2018). In recent years, some researchers have applied the appraisal theory (Martin 2000; Martin and White 2007) to conduct attitude analysis in texts (Puspita and Pranoto 2021; Li 2016). In translation research, the shifts of attitude between source text (ST) and target text (TT) are also gaining increasing attention and are often considered critical points in a translator's decision-making process (Munday 2012). According to Martin and White (2007), the appraisal framework consists of three interacting systems, i.e., engagement, attitude, and graduation. The attitude system concerns feelings, and comprises affect, judgment, and appreciation. There is no doubt about the applicability of appraisal theory to study attitudinal shifts in translation (Munday 2015), and some researchers have yielded many findings (e.g., Pan 2015; Zhao and Li 2021). Munday (2015) examined the application of appraisal theory to translation analysis and explored the potential for using engagement resources and graduation to determine translator/interpreter positioning. However, most of these studies implemented descriptive qualitative methods to uncover and describe the appraisal devices embedded in the text. Few researchers used corpus tools, such as Antconc, to aid text analysis based on appraisal theory (Lin et al. 2019). No previous study has investigated attitudinal shifts in self-translation. Since no corpus tool can fully support automatic text analysis based on appraisal theory, it would be harder for researchers to apply this theory in the analysis of longer texts, not to mention certain subtle lexical patterns dispersed throughout them.

The rapid development of corpus linguistics after the 1980s refers to language research based on applying corpus techniques and tools. Corpus-assisted discourse analysis has achieved satisfactory results (e.g., Bednarek and Caple 2014; Samaie and Malmir 2017; Matthews 2021). Mona Baker (1993; 1995) showed how corpus research in translation studies could be done and paved the way for corpus-based translation studies. Translation scholars have yielded fruitful findings in this field in the past 20 years (e.g., Olohan 2004; Baños, Bruti, and Zanotti 2013; Kim 2017; Cattoni et al. 2021;).

*Linguistic Inquiry and Word Count* (LIWC) is a powerful commonly-used text analysis program to study word use in social/personality psychology from the perspective of word frequency. This quantitative analysis tool was developed by James Pennebaker and his colleagues (Pennebaker, Francis, and Booth 2001). LIWC can be used to analyze texts of several different languages (including Chinese) by choosing corresponding inbuilt dictionaries. The text analysis module then compares each word in the text against a user-defined dictionary, and then the dictionary identifies which words are associated with which psychologically relevant category (O'Dea et al. 2017). Psychologists have used LIWC to conduct studies in different fields (Wu et al. 2021). Its accuracy and efficiency have been proven in different studies (e.g., del Pilar Salas-Zárate et al. 2014; O'Dea et al. 2017). Holtzman et al. (2019) used LIWC to identify linguistic markers of narcissism and showed that narcissism has significant associations with select LIWC word categories, such as more sexual words, more swear words, and fewer tentative words.

Lexical repetition, an essential cohesive strategy, is an important stylistic device. Stylistic analysis is a prerequisite for a successful translation (Čermáková 2015). Linguistic features of higher-than-normal frequency can be identified to support stylistic analysis. However, Mahlberg (2007) commented that "a human observer is not always aware of all the features that are important to a text. Although a translator may not see difficulties, there may still be subtle lexical relationships in a text that effect the presentation of information". The underlying assumption is that certain repetitive features in texts, especially long texts, are not necessarily easily noticeable. How do translators identify these features? The answer is corpus linguistics. "n-gram", a term in corpus linguistics, is one of the essential linguistic features that construct linguistic style, where the *n* specifies the number of consecutive words that are repeated. The other two terms that are sometimes used interchangeably with "n-gram" are "cluster" (Čermáková 2015) and "lexical bundle" (Biber et al. 1999).

Different people have different idiolectal preferences. Utilizing n-gram has been proven as an effective means to identify authors (e.g., Houvardas and Stamatatos 2006; Anwar, Bajwa, and Ramzan 2019). Wright (2017) applied a corpus linguistic approach to testing the accuracy of word n-grams in identifying the authors of anonymized email samples, with results showing that word n-grams between 2 and 6 words are most promising in authorship attribution. Therefore, lexical repetitions in the original text can possess distinctive features. Čermáková (2015) pointed out that identification and treatment of lexical repetition, an essential rhetorical device, in the source text are crucial in translation and that the application of corpus tools may help translators to uncover subtle patterns in a text by identifying particular elusive n-grams and keywords to keep their translation consistent. Since lexical repetitions in different chapters are often made unconsciously, the modification of which in self-translation may reflect attitudinal changes in self-translators. If significant attitudinal changes exist, we can use corpus tools to find these modifications in the self-translation by searching for different translations of the original n-grams containing *attitude* words.

Lin's *Between Tears and Laughter* is for foreign readers, and its Chinese translation is for Chinese readers, each having slightly different writing purposes. The intent to persuade other people spontaneously increases the emotionality of individuals' appeals via their words (Rocklage, Rucker, and Nordgren 2018). The Chinese version does not need to "persuade" domestic readers to trust Chinese culture. Therefore, as a self-translator with greater freedom of revision and improvement of the original text, did Lin make some changes in the self-translation of this book to decrease emotionality significantly owing to different readerships and writing purposes?

This study used corpus tools to analyze the possible attitudinal changes in Lin Yutang's self-translation of *Between Tears and Laughter*. Since lexical repetition is an important stylistic device, it is essential that the co-translator keep the translation consistent in style. The main goal of the current study was to address the following two research questions: (1) Compared with the original work, are self-translation's different writing purposes and readerships likely to cause the self-translator to have attitudinal changes, which might be reflected by certain changes to the original n-grams in the source text? (2) Can the application of corpus tools help to uncover attitudinal changes of the self-translator in self-translation and aid the co-translator to translate in a similar style?

**Method and Data Analysis**

This study aimed to select one or two attitudes reflected in the text to study Lin's attitudinal changes in self-translation from the perspective of psycholinguistics by using LIWC2015 (James W Pennebaker et al. 2015) and AntConc3.5.8. Firstly, this study selected the first 11 chapters of *Between Tears and Laughter*, published by John Day Book Company, to build Corpus A, and chose the corresponding Chinese chapters published by Northeast Normal University Press to build Corpus B. Corpus A contains 31020 words, while Corpus B consists of 29784 words. Each corpus comprises 11 texts, representing the 11 chapters. Secondly, this study selected the last 12 chapters of the English and Chinese versions, respectively, to build Corpus C and Corpus D. There are 38742 words in Corpus C and 33678 words in Corpus D. This study used the OCR function of Adobe Acrobat to extract text from the two versions. All the Chinese characters were segmented by ICTCLAS2012 developed from ICTCLAS 2003 (Zhang et al. 2003). All the corpora underwent strict manual checking, and some contents irrelevant to the study, such as prefaces, annotations, and postscripts, were removed. This study took the following seven steps to conduct data analysis and only chose the important results to be shown in the following five tables due to space restrictions.

First, using the corresponding dictionaries of LIWC2015 to analyze Corpus A and Corpus B, respectively (see Table 1).

Second, calculating the mean, standard deviation (SD), median, and interquartile range (IQR) values of each word category (see Table 2).

Third, comparing Corpus A and Corpus B about all of the variables of psychological processes. "*t*-test" or "Mann-Whitney U Test" was used depending on whether or not the normality assumption and equality of variance were met (see Table 3).

Fourth, selecting the word categories representing the author's attitudes from the comparison groups with significant differences and using LIWC2015 to obtain all the words in the corresponding word categories. Only the top 10 words of relevant word categories are shown in Table 4.

Fifth, using AntConc3.5.8 to process both corpora to find respectively word n-grams of between 3 and 7 words in length, whose minimum frequency is 3. Next, n-grams containing words representing the author's attitudes were obtained.

Sixth, using Corpus A and B to build a parallel corpus aligned at the sentence level. Comparing the source text and Chinese translation to examine the author's attitudinal changes.

Seventh, repeating some of the six steps to analyze one or two attitudes in Corpus C and Corpus D (see Table 5).

**Results**

LIWC2015 classifies Psychological Processes into nine main word categories, i.e., drives, time orientations, relativity, personal concerns, affective, social, cognitive, perceptual, and biological processes (James W Pennebaker et al. 2015). These nine categories have 39 sub-categories. Since each main category is composed of several sub-categories, significant differences in sub-categories will not necessarily cause significant differences in main categories. After comparing all the frequencies of the nine main categories and 39 sub-categories of Corpus A with those of Corpus B, the results show that the overall difference is not marked, proving that Lin's self-translation basically follows the principle of translation equivalence. Only three main categories have significant differences, while 16 sub-categories show significant differences, several of which are caused by linguistic differences. For example, the significant difference in the sub-category "male references" is because pronouns are used more often in English. The significant differences in all three sub-categories of the main category "time orientations" is because English pays more attention to the use of tenses than Chinese.

This study focused on the word categories that can reflect the self-translator's attitudinal changes. According to the statistical comparison results, the following categories are worthy of attention. The main category of "affective processes" is composed of the sub-category "positive emotion" and "negative emotion". The sub-category "negative emotion" is further divided into "anxiety", "anger", and "sadness". Only the sub-category "anger" has a significant difference. However, the statistic of this sub-category alone leads to the significant difference in "negative emotion", which in turn makes a significant difference in "affective processes".

The main category of "drives" is composed of sub-categories "affiliation", "achievement", "power", "reward", and "risk". The main category "drives" and sub-categories "achievement" and "power" show no significant differences, while sub-categories "affiliation", "reward", and "risk" all have significant differences.

The results of the fourth step show that the top 3 words of the sub-category "affiliation" in Corpus A are "we", "our", and "us", the word counts of which are respectively 266, 95, and 45. In contrast, the top 3 words in Corpus B are "我们"("we"), "家"("home"), and "社会"("society"), with their respective word counts being 212, 30, and 30. The difference is partly because pronouns are used more often in English than in Chinese. Moreover, there are only two occurrences of n-grams worthy of attention: "our conception of" and "with her Allies".

The sub-category of "reward" has a significant difference. However, the results of the fifth step show that only the Chinese four-gram "如何 是 好"("what to do") is found, which occurs three times in Corpus B. For the relevant English n-grams, only four are found because they contain either "good" or "great", both of which are marked as category "reward" by LIWC. The three-gram "good manners and" occurs six times, while "as good as", "by good manners", and "a great nation" all appear three times. As for the three-gram "good manners and", Lin's translations for "good manners" are "礼, 礼貌, 礼让, 谦恭"(all mean "courtesy"), none of which is identified as Chinese category "reward". The other three-grams all show the same results. 73.33 percent of all the 4 English three-grams appear in Chapters 8 and 10. Apart from this, Chapter 8's and 10's category "reward" frequencies are the top 2 among all 11 chapters. Both the chapters' main contents are to introduce traditional Chinese philosophical concepts. The significantly higher mean frequency of Corpus A in the category "reward" than that of Corpus B is in line with Lin's writing purpose that Western readers would benefit from learning and applying the good merits of Chinese philosophy. However, Chinese and English n-grams containing "reward" words are few, and their respective translations contain no words of the category "reward". Moreover, in those translations, no modifications of the original contents are found.

Although the sub-category "risk" has a significant difference($t$=-2.661, $p$=0.015), the results of the fourth step show that the top 3 words of this sub-category in Corpus B are "问题", "没有", and "相信"("believe in"), which are common words in Chinese, and in the most cases of this corpus do not mean real risks. "问题" can mean "question", "issue", "problem", "matter", "trouble", "difficulty", "key", and "defect". "没有" means "there is not" or "not have", but it is often used in double negatives in Chinese. The word count of these three words is 123 in total, taking up 38.32 percent of all the words in this category, which can make a significant difference insignificant.

In light of the above analysis, there may be words or phrases in the self-translation that can show Lin's attitudinal changes. However, there are few n-grams worthy of analyzing for categories "affiliation", "reward" and "risk". This paper focused on n-grams; thus, only the sub-category "anger" needed to be explored in depth. Parts of the results of the fourth step are shown in Table 4. According to Mona Baker (1995), a word is a token in a given corpus. The number of tokens is the number of words. The number of types is the number of different words. The results of the fifth step show that there are 639 types and 2486 tokens of English n-grams of between 3 and 7 words in length in Corpus A, while only 182 types and 674 tokens of Chinese n-grams of the same range of length are in existence in Corpus B. Only word n-grams containing "anger" words are selected, including 50 types of English n-grams of "anger" in Corpus A, 14 types of Chinese n-grams of "anger" in Corpus B. However, there may be overlaps between word n-grams of different lengths. For example, in the list of the 50 types of English n-grams of "anger", there are "the war to", "the war to end", "the war to end all", and "the war to end all wars", all of whose number of tokens are all 3.

**Discussion**

According to Table 1, the frequency difference of the sub-category "anger" between Corpus A and Corpus B in Chapter 10 is 0.67, which is the highest of all the 11 chapters. Due to this chapter's substantial attitudinal changes, the original contents may be modified in the Chinese self-translation, including words, phrases, or even whole sentences. Examples (1) to (3) fully demonstrate Lin's attitude changes in the self-translation owing to different writing purposes and readerships.

In Example (1), Lin translates the original "I am all for psychology" into "我并非反对心理学" ("I do not disapprove of psychology"). Since psychology studies originated in the West, the original text can shorten the distance with Western readers first and then promote traditional Chinese culture, which is more acceptable to the readers. There is a big difference between the translation and the original. In his self-translation, Lin looks at the problem from the perspective of domestic readers and is cautious about Western psychology.

(1) ST: **I am all for psychology**, because it alone holds the key to human behavior, … (Chapter 10)

我并非反对心理学，因人类行为必基于心理，...

In the English version, the source texts of Examples (2) and (3) are three sentences apart. In this context, Lin expresses dissatisfaction with the little support of the Allies to China. Moreover, he deems China was too courteous in foreign affairs and should be more assertive to protect its own interests. In Example (2), Lin only translates the word "bowing" and repeats it three times, omitting all the other words in this sentence. In Example (3), Lin not only omits the first English sentence, but also adds "中国新入强国之列" ("China has become a new member of world powers") in the translation. The reason for these modifications may be that both "looking pleased" and "kowtowing" seem too discouraging to domestic readers, which is not in line with the purpose of the self-translation. Lin intended to raise fighting spirit among Chinese people with this book. In contrast, it is better to point out the problems directly to Western readers to gain more attention. Thus, we can see Lin's different attitudes toward different readerships.

(2) ST: By China bowing **and looking extremely well pleased**. (Chapter 10)

叩头叩头又叩头。

(3) ST: **By our kowtowing we were misunderstood**. China has been acting like a college freshman just initiated into a fraternity, ... (Chapter 10)

中国新入强国之列，就像大学一年级新生，选入学生会馆...

The above three examples prove that self-translators' modifications of the original text can reveal attitudinal changes, as are the following n-grams, which occur at least three times. Nevertheless, the results of the sixth step show that modification in the self-translation only occurs once for each type of the following n-grams. The modification methods include omission and replacement. The rest of the n-grams all have the same or similar translations. Hence, the following examples of n-grams are unique and can reflect Lin's attitudinal changes.

### English N-grams of Word Category Anger

As mentioned in the introduction, this book's writing purpose differs from its self-translation purpose. Anger is a valuable means to impact and change the behaviors of others (van Kleef 2009). It is reasonable to show more anger in the English version than in the Chinese self-translation to show dissatisfaction and ask for more respect and assistance from the Allies. The words "war", "contempt", "bitter", and "烂"("rubbish") in the following n-grams are marked as category "anger" by the LIWC program.

*In this war (7 occurrences)*

In Example (4), combined with the context analysis, the original English sentence means that everything has a cause and effect, and that a country that wantonly provokes war does the opposite and will not escape the punishment of karma. The previous sentence of this one has the phrase "invisible forces" referring to the karmatic forces, which is translated into "潜势力" by Lin. "潜势力" is the Chinese equivalent of "invisible forces". The word "forces" in this example refers to the same concept. The word "war" in the original sentence refers to World War II, but the three-gram "in this war" is omitted in the Chinese translation. Because Lin points out in the book that China was passively dragged into the world war, and the cause of the war was the attempt of Western countries to rule the world by the ruling elite. China was invaded and forced to become one of the main battlefields of World War II. In a way, China was a victim. The Chinese version of the book was translated for domestic readers to expose the cruelty and inhumanity of the war and encourage Chinese people to be self-reliant rather than to increase sorrow. Therefore, the three-gram is omitted here.

In Example (5), Lin holds that government by courtesy and good manners or government by "rituals" should be implemented. The world should build lasting peace through cooperation based on good manners and mutual respect. However, western powers mistook China's courtesy and self-restraint for a sign of weakness. Lin argues that the United Kingdom and the United States were disrespectful at the Casablanca Conference in 1943 to decide the role of China in the war and hand it over to China for implementation. In the translation, Lin also omits "in this war" to create ambiguity so that readers can only speculate from the context in which event China played a subordinate role, so as to blur the cruelty of the war and China's helplessness. In Example (6), even though it is about America's pain, it will also make Chinese readers feel unwell. Hence, both "militarily" and "of the war" are omitted in the Chinese translation to make the hard fact less cruel and more acceptable to readers.

(4) ST: Politically, we ignore them. We are acting **in this war** as if these forces did not exist. (Chapter 3)

Lin: 政治上我们却置之不理，倒行逆施，仿佛没有这种潜势力存在。

(5) ST: China's role **in this war** in 1943, we are told, was discussed, decided upon, and handed out to China. (Chapter 10)

Lin: 据说，1943年中国应派何种工作，由他们讨论过，决定过，交给中国去奉行。

*Of the war (7 occurrences)*

(6) ST: ... he could be almost certain that the moral effect was as disastrous for Japan as the physical effect was **militarily** advantageous for her at the initial stage **of the war**. (Chapter 2)

Lin: ... 他可以断定这次袭击在精神上大不利于日人，和其物质上初期大利于日人相等。

*Contempt for the (5 occurrences)*

In Example (7), the Chinese equivalent for "contempt" should be "鄙夷". However, Lin translates "contempt" into "不重视"("neglect"). Both SnowNLP and BaiduSenta sentiment analysis were run to calculate the negative probabilities of "鄙夷" and "不重视". The BaiduSenta shows the negative probability of "鄙夷" is 0.8802, and that of "不重视" is 0.8766. The SnowNLP calculates "鄙夷" as 0.9247 and "不重视" as 0.7458. The results of both tools all prove that "鄙夷" is more negative than "不重视". The English sentence shows Lin does not agree with or even rejects Western mathematical thinking habits, while the Chinese translation reveals his strong identification with Chinese culture and more tolerance for Western culture. The word "中国"("China"), which is not in the original, is added in the translation, further confirming Lin's sense of cultural identity.

(7) ST: It is an unmathematical way of life, arising from **contempt for the** mathematical way of thinking. (Chapter 8)

Lin: 中国生活乃**不重**数学的生活，由于不重数学的思想习惯所造成。

*The bitter truth (4 occurrences)*

Example (8) is the 17th sentence behind Example (5). Examples (8) and (9) are two adjacent sentences, which mean that China does not need to remain humble all the time but should be moderately tough so as not to be mistaken as a weak country by Western countries. In the original English sentences, the two phrases, "the bitter truth", are adjacent to each other, separated by a period. The modifiers before the two truths are both "bitter". It expresses Lin's uncompromising attitude and anger. In Corpus A, "the bitter truth" appears four times. The other two occurrences of this three-gram refer to historical revolutions of Western countries. In the Chinese self-translation of Examples (8) and (9), Lin translates the first "bitter" into "逆耳的" and omits the second "bitter". The word "bitter" here means "making you feel very unhappy; caused by great unhappiness", while "逆耳的" means "unpleasant to the ear". There is less anger in the Chinese translation than in the English sentence. The original English version is written to convey anger to Western readers and achieve the purpose of criticizing the Allies. The Chinese self-translation is for domestic readers. Reducing anger can reduce the anxiety of domestic readers and boost our nation's self-confidence.

(8) ST: ... some Chinese must tell them **the bitter truth**. (Chapter 10)

Lin: ... 应当有一两个中国人告诉他们**逆耳的实话**。

(9) ST: **The bitter truth** is that behind the courteous front, resentment against the conduct of certain governments is very bitter, that the Chinese are frankly disappointed in their Allies, ... (Chapter 10)

Lin: 实话是，中国外貌非常客气，心中却是非常不满某国政府之行动。中国老老实实对于同盟失望，...

### *Chinese N-grams of Word Category Anger*

According to the statistics, there is no denying that Lin expresses less anger in the Chinese version. However, it is still necessary for Lin to express anger once China's core interests are violated. Example (10) shows us there is more anger in the self-translation than in the original text.

*汽油烂铁 (3 occurrences)*

In Corpus A, there are two occurrences of the four-gram "oil and scrap iron" and one occurrence of "scrap iron and oil". The first "oil and scrap iron" is in the first chapter, and the second one in Example (10) is in the tenth chapter. The "scrap iron and oil" is also in the first chapter but is behind the first "oil and scrap iron". China and America were members of the Allies in the second world war. The four-gram "scrap iron and oil" appears in the book to suggest humorously that China could also send strategic materials to the fighting enemies of China's "friends" as was done by America. Both occurrences of "oil and scrap iron" are about America's shipping strategic materials to Japan. However, Lin only points out America when this four-gram first appears in Chapter 1, and Chapter 1 and Chapter 10 are not close. When Western readers read this four-gram of Example (10) for the second time in Chapter 10, they may be unable to know exactly, even from the context of the English original, which country shipped oil and scrap iron to Japan since there is not only America in the context but also other two countries, i.e. Britain and Russia. Besides, it can be any other country since the context provides little crucial information.

Example (10) is closely behind Example (5) in Chapter 10. The word "scrap" in Example (10) means "things that are not wanted or cannot be used for their original purpose, but which have some value for the material they are made of". Therefore, the Chinese equivalent for "scrap iron" should be "废铁". The Chinese word "废" means

"useless or disused". The word "烂" used by Lin in this translation means "rotten or rubbish", which might have more anger. In Corpus B, all the above 2 English four-grams are translated into "汽油 烂 铁"("oil and rubbish iron"). Lin's anger lies in that America sold strategic materials to Japan, which could be used to make bombs to invade China. Moreover, "美国"("America") is added in the translation of Example (10), which will rouse Chinese readers' anger even more. Once Chinese readers are clear about which country did this, there is no reason not to vent their anger.

(10) ST: Silence and absence of protests against the shipping of **oil and scrap iron** to Japan were taken to mean profound satisfaction with the state of affairs. (Chapter 10)

Lin: 美国运**汽油烂铁**与日本，中国未抗议，被人误解以为十分感激满意。

**Attitude "Anger" in Xu Chengbin's Co-translation**

The results of the seventh step show that 7 out of the nine main word categories of Psychological Processes and 14 out of the 39 sub-categories have significant differences. Compared with the original text, one of the characteristics of Lin's self-translation is that there is less anger. One of the criteria to evaluate the quality of co-translation is that the co-translator can imitate the style of the principal translator. In this case, it is essential that Xu also contain the anger of the original text in the co-translation.

According to Table 5, the mean value of Corpus C's 12 frequencies is 1.13, while that of Corpus D's is 0.85. However, the results of a $t$-test of the two groups of frequencies in Table 5 show no significant difference between Corpus C and Corpus D in the category "anger" ($t$=1.88, $p$=0.07). We can see that although there is less anger in Xu's translation than in the corresponding part of Lin's original text, the difference is not that large. Xu does not contain the anger of the original text as much as Lin does. Since the occurrence frequency of the word "war" is the highest in Corpus A, in order to ensure comparability, any occurrences of n-grams containing the word "war" in Corpus C were found to check whether they have experienced some changes or not in their corresponding translations.

The results show that there is only one occurrence for each of the following three types of n-grams that undergo modifications in Corpus D, although each n-gram appears at least three times in Corpus C. The three n-grams are Chapter 18's "after the war", and Chapter 12's "in this war" and "of this war". In these three occurrences of modifications, Xu uses the same method of omission to omit the n-grams. The reason for these omissions may be that it is better to mention less about the war which brought nothing but agony to common Chinese people.

**Use of Corpus Tools to Aid Co-translation**

When Lin and Xu collaborated in translating *Between Tears and Laughter*, there were no corpus tools. Nowadays, self-translators and co-translators can use corpus tools such as LIWC and AntConc to make co-translation more efficient and successful.

When a co-translator has finished the translation of one or several paragraphs, he can run LIWC to check whether the differences in psychological processes between his translation and the corresponding part of the original text are close to those between the principal translator or self-translator's translation and the corresponding original part. If there are noticeable differences, it is convenient for the co-translator to use LIWC to see the psychological category of any word in the text so as to use proper translation methods to make some adjustments. It is necessary to exclude those word categories whose outcomes can be affected by linguistic differences between different languages.

Moreover, lexical repetition is not only an important stylistic device but also possesses distinctive features. By using AntConc and LIWC in synergy to analyze how the principal translator translates his own n-grams of the original text, especially those n-grams containing the words identified by LIWC that can reveal psychological states, the co-translator can make attempts to translate in a similar style.

**Conclusion**

The results of this study have provided answers to our two research questions. First, self-translators may undergo attitudinal changes when translating their own works for the readers of the target language due to different writing purposes. Some of the attitudinal changes can be identified by analyzing the modifications of the original text in the self-translation. Repetitive patterns that are not necessarily easily noticeable, especially those dispersed throughout a text, can reflect the author's attitudes. In many cases, authors of long texts are often unaware of such repetitive patterns. Self-translators' modifications of such lexical patterns can be efficiently identified by the text analysis tools LIWC and Antconc. We can use these two programs in synergy to check whether or not the n-grams containing attitude words experience modifications of psychological processes. This paper only focused on n-grams of attitude "anger" that abound in the text, which are also immune to the influence of linguistic differences between English and Chinese. *Between Tears and Laughter*, as shown above, may be an extreme case in which "anger" is an emotional attitude throughout the text. The findings show that there is less anger in the self-translation than in the English original due to different writing purposes and readerships. However, this study's approach may not work equally well for less expressive texts. Second, LIWC and Antconc have been proven to be efficient in identifying self-translators' attitudinal changes. However, the present study only points out the applicability of using corpus tools in this way to support co-translation. Further work in this direction is needed to develop a corpus-based methodology to support co-translation.

| Table 1 | | Frequencies of each word category in Corpus A and Corpus B | | | | |
|---|---|---|---|---|---|---|
| Corpora | | Corpus A | | | Corpus B | |
| Chapters | anger | affiliation | reward | anger | affiliation | reward |
| 1 | 1.43 | 1.58 | 1.19 | 0.89 | 0.99 | 0.83 |
| 2 | 0.98 | 3.80 | 0.87 | 0.63 | 2.16 | 0.48 |
| 3 | 0.96 | 1.87 | 0.68 | 0.46 | 1.82 | 0.57 |
| 4 | 1.08 | 2.93 | 0.84 | 1.04 | 2.54 | 0.50 |
| 5 | 1.61 | 2.25 | 0.88 | 1.06 | 1.85 | 0.58 |
| 6 | 1.65 | 2.20 | 0.73 | 1.05 | 1.76 | 0.52 |
| 7 | 0.58 | 1.56 | 0.89 | 0.85 | 1.32 | 0.81 |
| 8 | 0.85 | 1.77 | 1.57 | 0.25 | 1.27 | 0.34 |
| 9 | 0.72 | 1.87 | 0.52 | 0.48 | 1.20 | 0.40 |
| 10 | 1.53 | 2.43 | 1.70 | 0.86 | 1.60 | 0.31 |
| 11 | 0.75 | 1.49 | 0.84 | 0.96 | 0.91 | 0.86 |

Table 2　Statistics of the frequencies of word category *anger* in Corpus A and Corpus B

| | Corpus A | Corpus B |
|---|---|---|
| Mean | 1.1036 | 0.7755 |
| SD | 0.3861 | 0.2775 |
| Median | 0.98 | 0.86 |
| IQR | 0.78 | 0.56 |

Table 3　independent sample *t*-test for differences between means of frequencies of word category *anger* in Corpus A and Corpus B

| Category | T-value | P-value | M_eng | M_chi | SD_eng | SD_chi |
|---|---|---|---|---|---|---|
| anger | 2.2892 | 0.0331 | 1.1036 | 0.7755 | 0.3861 | 0.2775 |

Table 4　Frequencies of word category *anger* in Corpus A and Corpus B

| Corpora | Corpus A | | Corpus B | |
|---|---|---|---|---|
| Ranking | anger | Wordcount | anger | Wordcount |
| 1 | war | 120 | 战争("war") | 61 |
| 2 | fighting | 33 | 反对("object") | 11 |
| 3 | wars | 11 | 否认("deny") | 9 |
| 4 | fight | 11 | 冲突("conflict") | 9 |

| 5 | domination | 8 | 烂("rot") | 7 |
| 6 | punishments | 7 | 侮辱("insult") | 6 |
| 7 | contempt | 7 | 杀("kill") | 6 |
| 8 | lies | 5 | 破坏("destroy") | 6 |
| 9 | rebellion | 5 | 强硬("tough") | 5 |
| 10 | bitter | 5 | 鄙("despise") | 5 |

Table 5　　Frequencies of word category *anger* in Corpus C and Corpus D

| Corpora | Corpus C | Corpus D |
| --- | --- | --- |
| Chapters | anger | anger |
| 12 | 1.19 | 0.76 |
| 13 | 1.43 | 1 |
| 14 | 0.98 | 0.94 |
| 15 | 1.79 | 1.37 |
| 16 | 1.9 | 1.06 |
| 17 | 1.01 | 0.65 |
| 18 | 1.44 | 1.1 |
| 19 | 0.92 | 0.56 |
| 20 | 1.13 | 1.07 |
| 21 | 0.64 | 0.56 |
| 22 | 0.46 | 0.6 |
| 23 | 0.71 | 0.54 |